\documentclass[conference,10pt]{IEEEtran}
\pdfoutput=1

\usepackage{hyperref}
\usepackage{graphicx}
\usepackage{url}
\usepackage{float}
\usepackage{amsmath}
\usepackage{subfig}
\graphicspath{{figures/}}

\usepackage{cite}
\usepackage{amsmath}
\usepackage{hyperref} 
\usepackage{amssymb}
\usepackage{graphicx}
\usepackage{subfig}
\usepackage{algorithm}
\usepackage{algorithmic}
\usepackage{array}
\usepackage{rotating}
\usepackage{float}
\usepackage{mathtools}
\usepackage{csquotes}
\usepackage{tabularx}
\usepackage{xcolor}
\usepackage{cancel}
\usepackage{multirow}
\usepackage[bottom]{footmisc}

\begin{document}

\title{Maximal Jacobian-based Saliency Map Attack}

\author{\IEEEauthorblockN{Rey Wiyatno and Anqi Xu}
\IEEEauthorblockA{Element AI\\
Montr\'eal, Canada}}

\maketitle

\section{Introduction}

The Jacobian-based Saliency Map Attack~\cite{DBLP:conf/eurosp/PapernotMJFCS16} is a family of adversarial attack methods~\cite{42503,43405} for fooling classification models, such as deep neural networks for image classification tasks.
By saturating a few pixels in a given image to their maximum or minimum values, JSMA can cause the model to misclassify the resulting \textit{adversarial image} as a specified erroneous \textit{target class}.
We propose two variants of JSMA, one which removes the requirement to specify a target class, and another that additionally does not need to specify whether to \textit{only} increase or decrease pixel intensities.
Our experiments highlight the competitive speeds and qualities of these variants when applied to datasets of hand-written digits and natural scenes.

\section{Jacobian-based Saliency Map Attack (JSMA)}

Saliency maps were originally conceived for visualizing the prediction process of classification models~\cite{DBLP:journals/corr/SimonyanVZ13}.
The map rates each input feature $x_{(i)}$ (e.g. each pixel)\footnote{We use the notation $v_{(i)}$ to denote the $i$-th element of the vector $v$.} on how influential it is for causing the model to predict a particular class $c = \hat{y}(x) = \arg \max_{c'} f(x)_{(c')}$, where $f(x)$ is the softmax probabilities vector predicted by the victim model.

One formulation of the saliency map is given as:

\begin{equation*} \label{equations:jsma_increase}
S^+(x_{(i)}, c) = 
\begin{cases}
 0~~~\text{if} \ \frac{\partial f(x)_{(c)}}{\partial x_{(i)}} < 0 \ \text{or} \ \sum\limits_{c' \neq c} \frac{\partial f(x)_{(c')}}{\partial x_{(i)}} > 0 \\
 - \frac{\partial f(x)_{(c)}}{\partial x_{(i)}} \cdot \sum\limits_{c' \neq c} \frac{\partial f(x)_{(c')}}{\partial x_{(i)}}~~~\text{otherwise}
 \end{cases}
\end{equation*}

\noindent $S^+(\cdot)$ measures how much $x_{(i)}$ \emph{positively} correlates with $c$, while also \emph{negatively} correlates with all other classes $c' \neq c$.
If either condition is violated, then saliency is reset to zero.

An attacker can exploit this saliency map by targeting an adversarial class $t$ that does not match the true class label $y$ of a given sample $x$.
By \textit{increasing} a few high-saliency pixels $x_{(i)}$ according to $S^+(x_{(i)}, c=t)$, the modified image $x'$ will have an increased prediction confidence $f(x')_{(t)}$ for the adversarial class $t \neq y$, and thus might result in misclassification.

Alternatively, one can attack by \emph{decreasing} feature values based on another saliency map $S^-(\cdot)$, which differs from $S^+$ only by the inversion of the low-saliency inequalities, i.e. $S^-\left(x_{(i)},t\right) = 0$ if $\frac{\partial f(x)_{(t)}}{\partial x_{(i)}} > 0 \ \text{or} \ \sum_{c \neq t} \frac{\partial f(x)_{(c)}}{\partial x_{(i)}} < 0$.

In practice, both saliency measures $S^+$ and $S^-$ are overly strict when applied to individual input features, because often the summed gradient contribution across all non-targeted classes $\sum_{c' \neq c} \frac{\partial f(x)_{(c')}}{\partial x_{(i)}}$ will trigger the minimal-saliency criterion.
Thus, the Jacobian-based Saliency Map Attack (JSMA)~\cite{DBLP:conf/eurosp/PapernotMJFCS16} alters these saliency measures to search over \emph{pairs} of pixels $(p, q)$ instead.

\begin{figure}[t]
    \centering
    \includegraphics[width=1.0\linewidth]{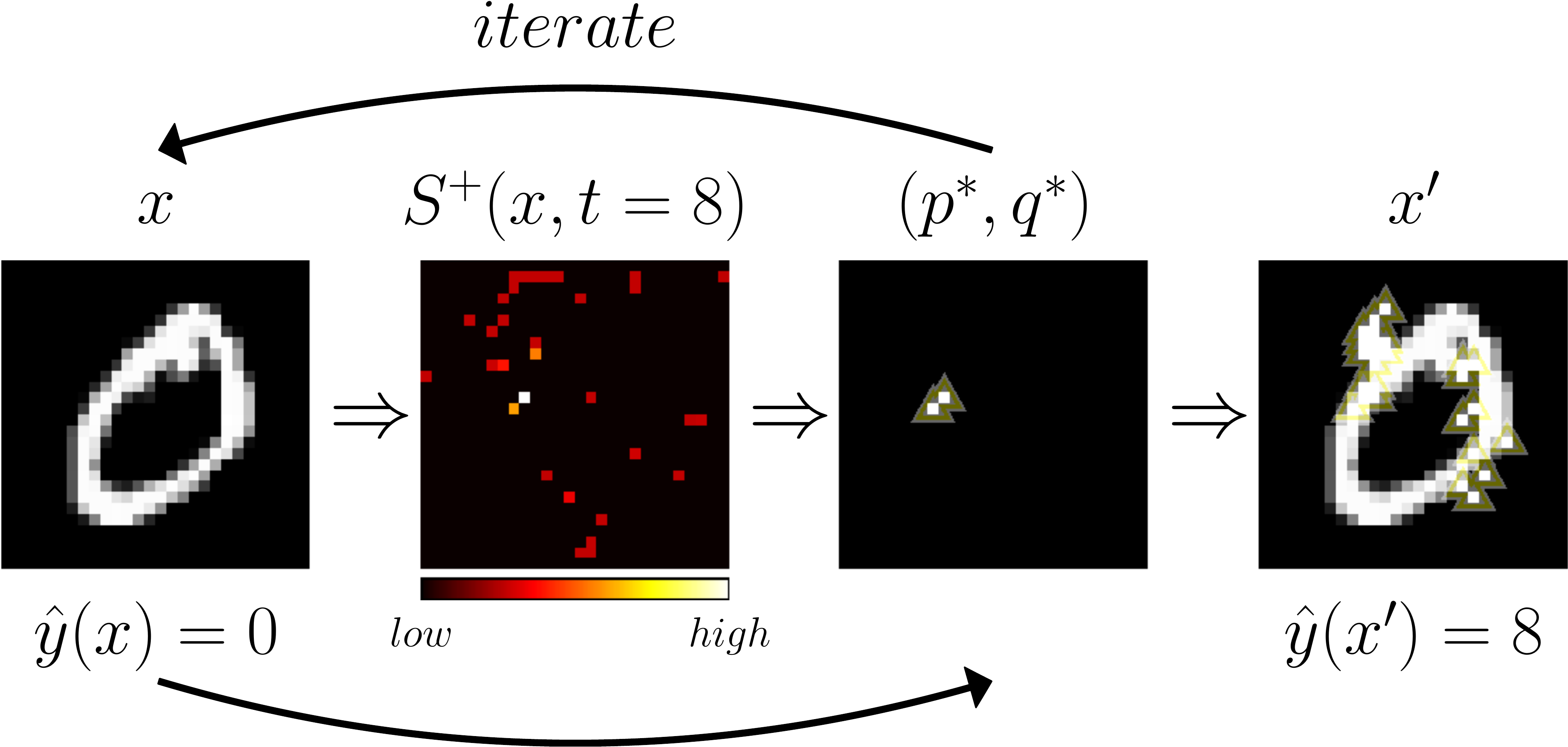}
    \caption{Illustration of JSMA+F algorithm.}
    \label{fig:jsma_plusF}
    \vspace{-0.3cm}
\end{figure}

Given a unit-normalized input $x$, JSMA initializes the search domain $\Gamma$ over all input indices. Then, as illustrated in Fig.~\ref{fig:jsma_plusF}, it finds the most salient pixel pair $(p^*,q^*)$, perturbs both values (by $\theta = +1$ if using $S^+$, or $\theta = -1$ if using $S^-$), and then removes saturated feature indices from $\Gamma$. This process is repeated until either the model misclassifies the perturbed input $x'$ as the target class $t$ or till a maximum number of iterations $I_{max}$ is reached.

Carlini and Wagner~\cite{DBLP:conf/sp/Carlini017} proposed an alternation that amplifies the logit $Z(x)_{(t)}$ rather than the softmax probability $f(x)_{(t)}$. We denote these variants as JSMA+Z, JSMA-Z, JSMA+F, and JSMA-F, based on the choices of increasing ($+$) or decreasing ($-$) feature values, and using $Z$ versus $f$.

The original authors advocated saturating perturbations $\theta=\pm 1$ to find $x'$ with the \textit{fewest} feature changes (i.e. minimal $L_0$ norm). In some domains such as hand-written digits, we anecdotally note that adversaries found with $\theta < 1$ had smaller $L_2$ perturbed distances and were more perceptually similar thus less likely to be detected by humans.

Furthermore, we suggest that the maximum per-feature perturbation ($L_{\infty}$ norm) can be optionally $Clip$-ped to within an $\epsilon$-neighborhood to further limit perceptual differences, similar to the BIM attack~\cite{DBLP:journals/corr/KurakinGB16}:

\vspace{-0.5cm}

\begin{equation*} \label{equations:clip}
\small
Clip_{\epsilon} \{x'_{(i)} \} =  \textrm{min} \Big\{ 1, x_{(i)} + \epsilon, \textrm{max} \Big\{ 0,x_{(i)} - \epsilon,x'_{(i)} \Big\} \Big\}
\end{equation*}



\section{Non-Targeted JSMA (NT-JSMA)}

All JSMA variants above must be given a specific target class $t$. This choice affects the speed and quality of the attack, since misclassification under certain classes are harder to attain than others, such as trying to modify a hand-written digit ``$1$'' to look like anything other than ``$7$''~\cite{DBLP:conf/eurosp/PapernotMJFCS16}.

\begin{figure}[t]
    \centering
    \includegraphics[width=1.0\linewidth]{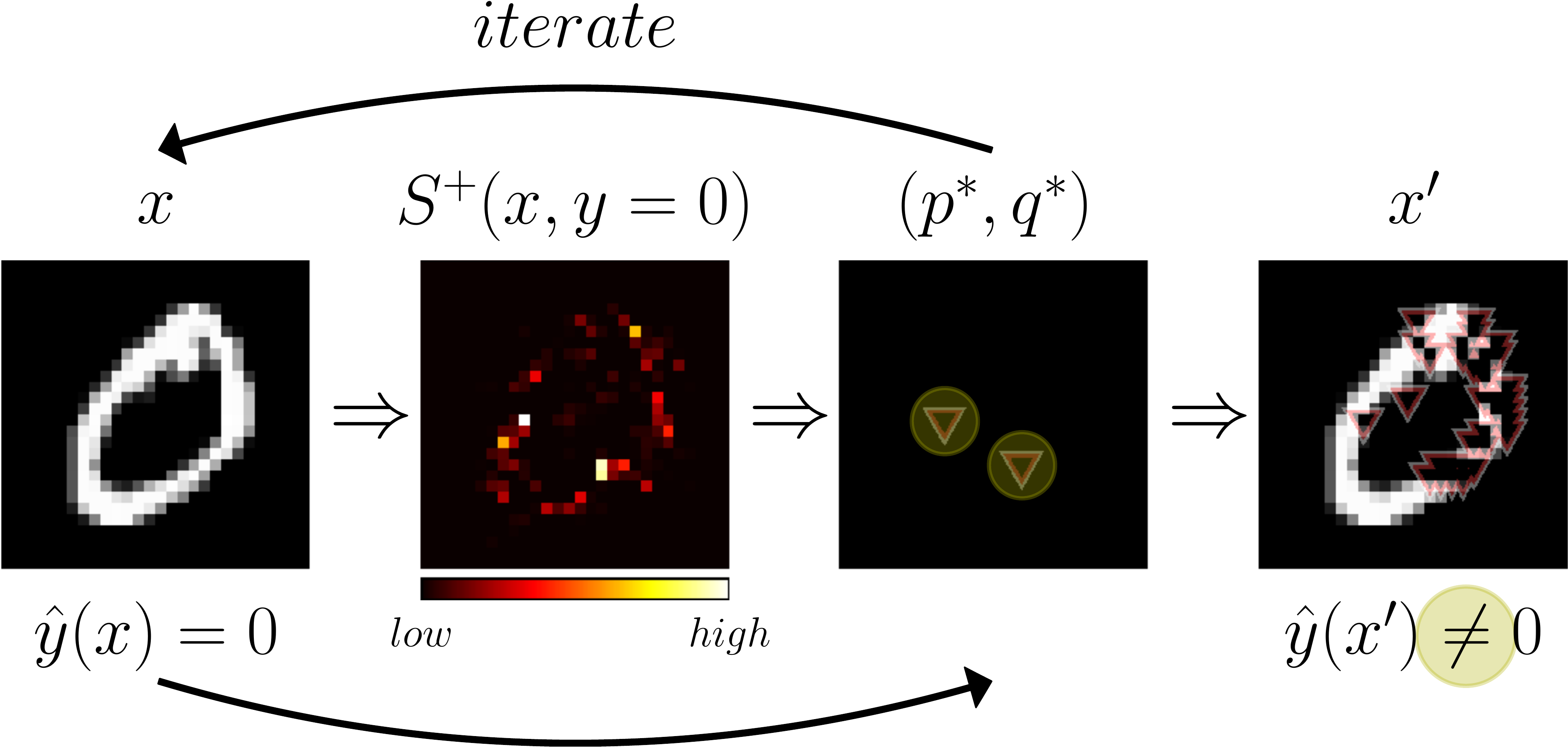}
    \caption{Illustration of NT-JSMA-F algorithm, with highlighted differences from JSMA+F.}
    \label{fig:ntjsma_minusF}
\end{figure}

We propose a non-targeted attack formulation that removes this target-class dependency by having the algorithm \textit{decrease} the model's prediction confidence of the \textit{true} class label ($c=y$), instead  of \textit{increasing} the prediction confidence of an adversarial target $t \neq y$. As depicted in Fig.~\ref{fig:ntjsma_minusF}, the NT-JSMA procedure is realized by swapping the saliency measure employed, i.e. following $S^-$ when increasing feature values (NT-JSMA+F / NT-JSMA+Z), or $S^+$ when decreasing feature values (NT-JSMA-F / NT-JSMA-Z). This variant also naturally relaxes the success criterion, such that an adversarial example $x'$ only needs to \textit{not} be classified as the true class $y$, i.e. $\arg \max_c f(x')_{(c)} \neq y$.

\section{Maximal JSMA (M-JSMA)}

\setlength{\textfloatsep}{8pt}
\begin{algorithm}[b]
   \caption{\textcolor{blue}{Maximal} Jacobian-based Saliency Map Attack (\textcolor{blue}{M-}JSMA\_F)}
   \label{alg:m-jsma}
\begin{algorithmic}
  \footnotesize
   \STATE {\bfseries Input:} $x \in [0,1]^n$, \textcolor{red}{\cancel{target class $t$,}} \textcolor{blue}{true class $y$}, classifier $f$, $I_{max}$, perturbation step $\theta$ \textcolor{red}{\cancel{$=+1$}} \textcolor{blue}{$\in (0, 1]$}, \textcolor{blue}{max. perturbation bound $\epsilon \in (0,1]$}
   \STATE {\bfseries Initialize:} $x' \leftarrow x$, $i \leftarrow 0$, $\Gamma = \{1,...,n\}$, \textcolor{blue}{$\eta = \{0,0,...\}^n$}
   \WHILE{$\hat{y}(x')$ \textcolor{red}{\cancel{$\neq t$}} \textcolor{blue}{$== y$} \AND $i < I_{max}$ \AND $|\Gamma| \geq 2$ }
   \STATE $\gamma \leftarrow 0$
   \FOR{every pixel pair ($p,q$) $\in \Gamma$ \textcolor{blue}{and every class $t$}}
   \STATE $\alpha \leftarrow \sum_{k = p,q} \frac{\partial f(x')_{(t)}}{\partial x'_{(k)}}$
   \STATE $\beta \leftarrow \sum_{k = p,q}\sum_{c \neq t} \frac{\partial f(x')_{(c)}}{\partial x'_{(k)}}$
   \IF {\textcolor{red}{\cancel{$\alpha>0 \ \AND \ \beta<0 \ \AND \ $}} $-\alpha \cdot \beta > \gamma$}
   \STATE $(p^*,q^*), \gamma \leftarrow (p,q), -\alpha \cdot \beta$
   \STATE \textcolor{blue}{
   $\theta' \leftarrow
   \begin{cases}
   - sign(\alpha) \cdot \theta & \text{if} \ t==y\\
   sign(\alpha) \cdot \theta & \text{otherwise}
   \end{cases}$
   }
   \ENDIF
   \ENDFOR
   \STATE $\textbf{if}~\gamma == 0~\textbf{then}~\textbf{break}$
   \STATE $x'_{(p^*)}, x'_{(q^*)} \leftarrow \textcolor{blue}{Clip_{\epsilon}\{}(x'_{(p^*)} + \textcolor{red}{\cancel{\theta}} \textcolor{blue}{\theta'})\textcolor{blue}{\}}, \textcolor{blue}{Clip_{\epsilon}\{}(x'_{(q^*)} + \textcolor{red}{\cancel{\theta}} \textcolor{blue}{\theta'})\textcolor{blue}{\}}$
   \STATE $\text{Remove} \ p^* \ \text{from} \ \Gamma \ \textbf{if} \ x'_{(p^*)} \notin (0, 1)~\textcolor{blue}{\textbf{or}~\eta_{(p^*)} == -\theta'}$
   \STATE $\text{Remove} \ q^* \ \text{from} \ \Gamma \ \textbf{if} \ x'_{(q^*)} \notin (0, 1)~\textcolor{blue}{\textbf{or}~\eta_{(q^*)} == -\theta'}$
   \STATE \textcolor{blue}{$\eta_{(p^*)}, \eta_{(q^*)} \leftarrow \theta'$}
   \STATE $i \leftarrow i + 1$
   \ENDWHILE
   \STATE {\bfseries Return:} $x'$
\end{algorithmic}
\end{algorithm}

In addition to alleviating the need to specify a target class $t$, we can further alleviate the need to specify whether to \textit{only} increase or decrease feature values. The resulting Maximal Jacobian-based Saliency Map Attack (M-JSMA) combines targeted and non-targeted strategies and considers \textit{both} the increase and decrease of feature values.
As shown in Algorithm~\ref{alg:m-jsma}\footnote{Algorithm~\ref{alg:m-jsma} contrasts with the original JSMA+F~\cite{DBLP:conf/eurosp/PapernotMJFCS16}, where \textcolor{blue}{blue text} denotes additions for M-JSMA and \textcolor{red}{\cancel{red text}} denotes omitted parts of JSMA+F.}, at each iteration the maximal-salient pixel pair is \textit{chosen over every possible class} $t$, whether adversarial or not. Also, instead of enforcing low-saliency conditions via $S^+$ or $S^-$, we simply identify the most salient pair $(p^*,q^*)$ according to either map, and consequently \textit{decide on the perturbation direction $\theta'$ accordingly}. An additional history vector $\eta$ is added to prevent oscillatory perturbations. As for NT-JSMA, M-JSMA terminates when the predicted class for $x'$ no longer matches the true class.

\section{Evaluation}

\begin{figure}[h]
    \centering
    \includegraphics[width=1.0\linewidth]{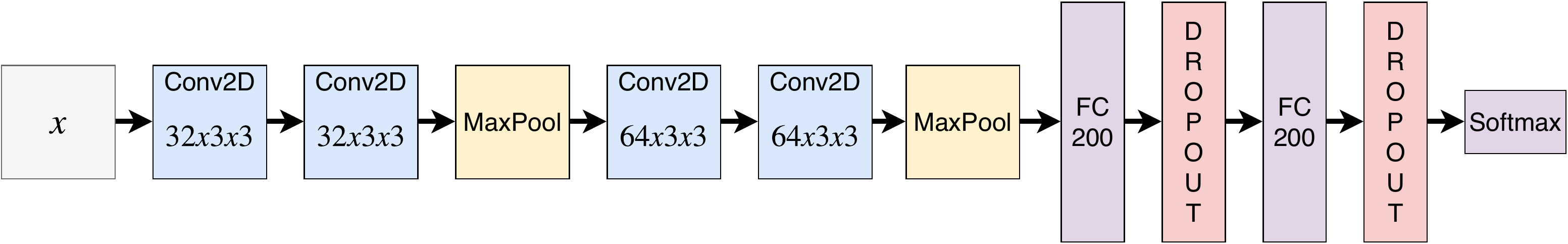}
    \caption{Architecture for baseline MNIST classifier model.~\cite{7546524}}
    \label{fig:mjsma_victim}
\end{figure}

We trained classifiers following the \textit{baseline MNIST} architecture introduced in~\cite{7546524}, which is depicted in Fig.~\ref{fig:mjsma_victim}.
The test-set accuracies of our baseline models for MNIST~\cite{726791}, Fashion-MNIST~\cite{DBLP:journals/corr/abs-1708-07747}, and CIFAR10~\cite{Krizhevsky2009LearningML} are $99.53\%$, $92.38\%$, and $84.53\%$, respectively.

\begin{figure*}[ht!]
    \centering
    \includegraphics[width=1.0\linewidth]{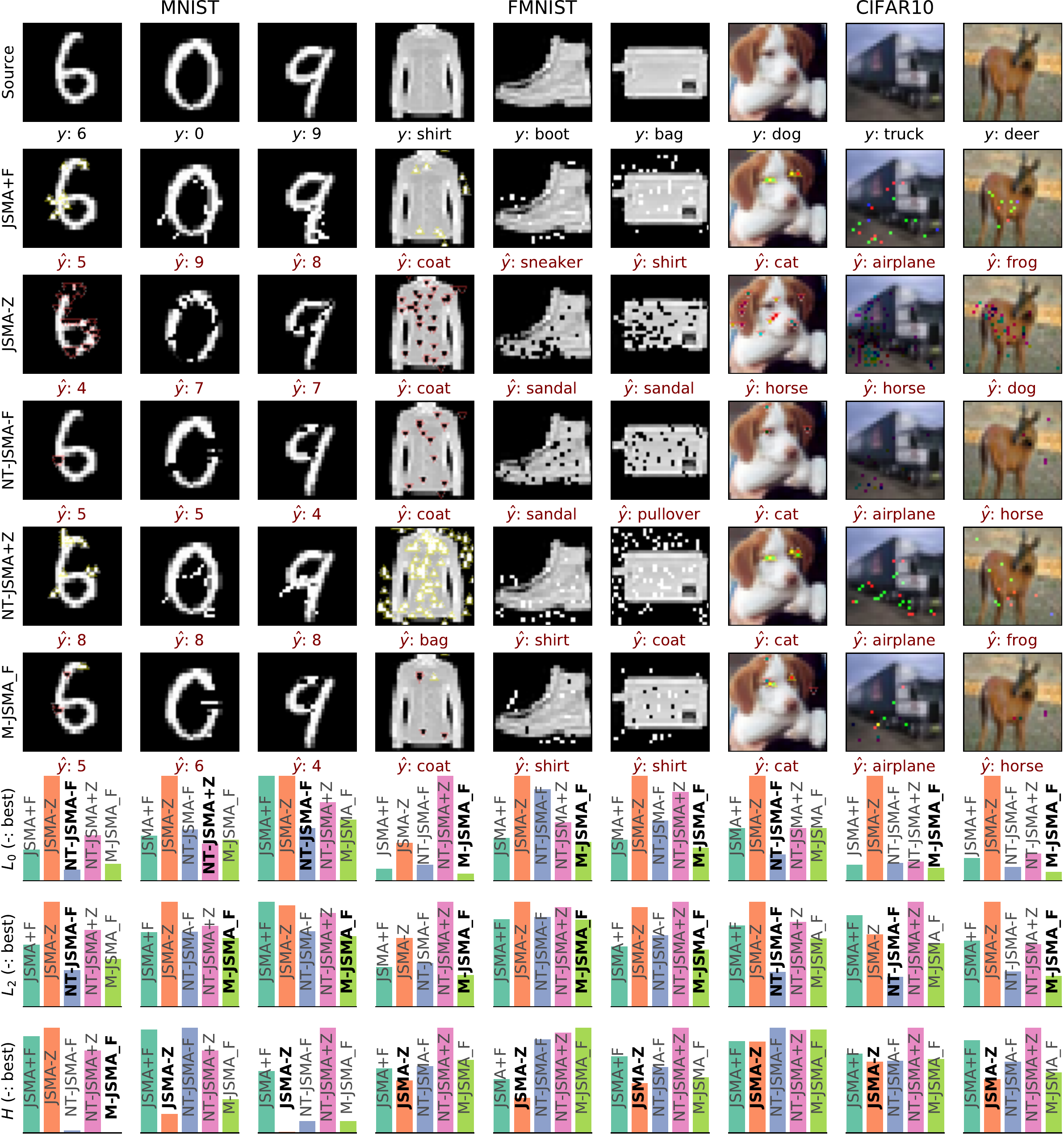}
    \caption{Samples of adversarial examples found by various JSMA variants, along with various evaluation metrics: $L_0$ for convergence speed, $L_0$ and $L_2$ for perceptual similarity, and softmax entropy $H$ for prediction uncertainty of the adversaries.}
    \label{fig:mjsma_samples}
\end{figure*}

We applied the various JSMA variants to all \textit{correctly-classified} test-set instances, using $I_{max}=\infty$.
Each model+dataset attack run is evaluated on its \textit{success rate} (\%), the average $L_0$ distance (which also reflects \textit{convergence speed} when $\theta=\pm 1$), the average $L_2$ \textit{perceptual distance}, and the average softmax entropy (H) reflecting \textit{misclassification uncertainty}.
To compare best-case performance, when evaluating targeted attacks on each sample, we focus on the \textit{single} target class that results in misclassification in the \textit{fewest} iterations possible.
Samples of adversarial examples are shown in Fig.~\ref{fig:mjsma_samples}.

\subsection{JSMA vs. NT-JSMA vs. M-JSMA}

\begin{table}[b]
\scriptsize
\setlength{\tabcolsep}{1.2pt}
\centering
\caption{Comparison of $\pm F$ variants ($|\theta|=1$, $\epsilon=1$).}
\begin{tabular}{l|rrrr|rrrr|rrrr}
\multirow{2}{*}{Attack} & \multicolumn{4}{c|}{MNIST} & \multicolumn{4}{c|}{F-MNIST} & \multicolumn{4}{c}{CIFAR10}\\
{} & \% & $L_0$ & $L_2$ & H & \% & $L_0$ & $L_2$ & H & \% & $L_0$ & $L_2$ & H \\
\hline
JSMA+F & 100 & 16.9 & 3.26 & \textbf{0.58} & 99.9 & \textbf{17.0} & 3.17 & 0.98 & 100 & \textbf{15.0} & 2.24 & 1.10 \\
\hline
JSMA-F & 100 & 18.5 & 3.30 & 0.62 & 99.9 & 31.1 & \textbf{2.89} & \textbf{0.90} & 100 & 16.6 & \textbf{1.59} & \textbf{1.05} \\
\hline
NT-JSMA+F & 100 & 17.6 & 3.35 & 0.64 & 100 & 18.8 & 3.27 & 1.03 & 99.9 & 17.5 & 2.36 & 1.16 \\
\hline
NT-JSMA-F & 100 & 19.7 & 3.44 & 0.70 & 99.9 & 33.2 & 2.99 & 0.98 & 99.9 & 19.6 & 1.68 & 1.12 \\
\hline
M-JSMA\_F & 100 & \textbf{14.9} & \textbf{3.04} & 0.62 & 99.9 & 18.7 & 3.42 & 1.02 & 99.9 & 17.4 & 2.16 & 1.12 \\
\hline
\end{tabular}
\label{table:exp1}
\end{table}

Looking at the average $L_0$ statistics reflecting perceptual similarity and convergence speed in Table~\ref{table:exp1}, we observe that across all 3 datasets it is consistently faster to find adversaries by \textit{increasing} pixel intensities rather than decreasing them. We also note that M-JSMA\_F found adversaries with similar number of pixel changes (and thus in similar number of iterations) compared to JSMA+F. On the other hand, the non-targeted variants consistently took one or two more iterations than their targeted counterparts.

Considering next the perceptual similarities as measured by average $L_2$ statistics, our results showed strong preferences for the pixel-decreasing variants, JSMA-F and NT-JSMA-F. This can be attributed to the fact that most images from the 3 datasets have dark backgrounds. Also, although adversaries found by M-JSMA\_F had the smallest perceptual similarity scores for MNIST, results for other datasets did not reflect similar benefits. Furthermore, we note again that NT-JSMA had slightly worse $L_2$ statistics compared to JSMA.

Finally, analyzing the uncertainties of adversarial predictions as reflected by average entropy $H$ statistics, we see that the original targeted JSMA formulations consistently found adversaries with lower-uncertainty predictions, especially compared to the non-targeted variants. Nevertheless, adversaries found by Maximal JSMA still showed competitive $H$ values on average.

Based on the results above, we conclude that the flexiblity of not specifying a target class in NT-JSMA resulted in minor added inefficiencies in terms of both convergence time and quality of adversaries. On the other hand, as M-JSMA considered all possible class targets, and both $S^+$ and $S^-$ metrics and perturbation directions, it inherited both the performance benefits and flexibilities among all other variants.

\subsection{Effects of Smaller Feature Perturbations}

\begin{table}[t]
\scriptsize
\setlength{\tabcolsep}{1.2pt}
\centering
\caption{Comparison of $\pm F$ variants ($|\theta|=0.1$, $\epsilon=0.5$).}
\begin{tabular}{l|rrrr|rrrr|rrrr}
\multirow{2}{*}{Attack} & \multicolumn{4}{c|}{MNIST} & \multicolumn{4}{c|}{F-MNIST} & \multicolumn{4}{c}{CIFAR10}\\
{} & \% & $L_0$ & $L_2$ & H & \% & $L_0$ & $L_2$ & H & \% & $L_0$ & $L_2$ & H \\
\hline
JSMA+F & 100 & 41.6 & 1.95 & \textbf{0.79} & 99.9 & 30.4 & 0.61 & 0.94 & 100 & 23.6 & 0.60 & 1.02 \\
\hline
JSMA-F & 80.3 & 44.7 & 2.24 & 0.81 & 99.2 & 48.5 & 1.16 & \textbf{0.89} & 100 & \textbf{22.8} & 0.58 & \textbf{1.01} \\
\hline
NT-JSMA+F & 99.9 & 36.5 & 1.93 & 0.86 & 99.5 & 32.5 & 0.64 & 1.01 & 99.3 & 26.3 & 0.63 & 1.12 \\
\hline
NT-JSMA-F & 54.2 & 34.6 & 1.98 & 0.89 & 94.5 & 48.6 & 1.15 & 0.95 & 96.2 & 24.9 & 0.59 & 1.11 \\
\hline
M-JSMA\_F & 98.2 & \textbf{31.5} & \textbf{1.71} & 0.85 & 99.5 & \textbf{30.3} & \textbf{0.60} & 0.98 & 98.4 & 23.3 & \textbf{0.54} & 1.08 \\
\hline
\end{tabular}
\label{table:exp2}
\end{table}

By perturbing features at increments of $|\theta|= 0.1$ and bounding $L_{\infty}$ to $\epsilon = 0.5$, Table~\ref{table:exp2} shows that all variants found adversaries with smaller perceptual differences (i.e. smaller $L_2$), albeit requiring more search time (i.e. larger $L_0$). Also, M-JSMA resulted in stellar convergence speeds and quality of adversaries compared to the other variants across datasets. Thus, we conclude that regardless of whether adversaries with fewer feature changes ($L_0$) or smaller Euclidean distances $(L_2$) are desirable according to a given application domain, M-JSMA performed favorably over other variants.

\subsection{Performance under Defensive Distillation}

Defensive distillation~\cite{7546524} is an adversarial defense method that re-trains a model using ground truth labels that are no longer one-hot-encoded, but rather using softmax probabilities resulting from the original model's logits divided by a temperature constant $T$. At inference time, $T$ is reset back to 1. As result, the gradients of the model approaches 0 as $T$ increases, which is a form of gradient masking~\cite{Papernot:2017:PBA:3052973.3053009,DBLP:journals/corr/PapernotMSW16,tramèr2018ensemble}. 

To test the effects of this adversarial defense strategy, we distilled our baseline models at $T=1$ (plain non-defensive distillation) and $T=100$. The resulting classifiers at $T=1$ on MNIST, F-MNIST, and CIFAR10 had test-set accuracies of 99.39\%, 91.92\%, and 83.19\%, respectively, while the accuracies for the distilled models at $T=100$ were 99.44\%, 91.81\%, and 83.55\%, respectively.

\begin{table}[t]
\scriptsize
\setlength{\tabcolsep}{1.0pt}
\centering
\caption{Comparison of $\{\pm F, \pm Z\}$ variants ($|\theta|=1$, $\epsilon=1$) on defensively distilled models with $T \in \{1,100\}$.}
\begin{tabular}{l|r|rrrr|rrrr|rrrr}
\multirow{2}{*}{Attack} & \multirow{2}{*}{$T$} & \multicolumn{4}{c|}{MNIST} & \multicolumn{4}{c|}{F-MNIST} & \multicolumn{4}{c}{CIFAR10}\\
{} & {} & \% & $L_0$ & $L_2$ & H & \% & $L_0$ & $L_2$ & H & \% & $L_0$ & $L_2$ & H \\
\hline
JSMA+F & 1 & 100 & 17.9 & 3.23 & \textbf{0.79} & 100 & \textbf{17.5} & 3.16 & 1.09 & 100 & \textbf{15.5} & 2.16 & 1.28 \\
\hline
JSMA-F & 1 & 100 & 16.9 & 3.18 & 0.84 & 100 & 32.3 & \textbf{2.96} & \textbf{1.00} & 100 & 19.3 & \textbf{1.84} & \textbf{1.24} \\
\hline
NT-JSMA+F & 1 & 100 & 18.8 & 3.31 & 0.84 & 100 & 18.9 & 3.25 & 1.13 & 100 & 18.6 & 2.27 & 1.34 \\
\hline
NT-JSMA-F & 1 & 100 & 17.8 & 3.28 & 0.93 & 100 & 34.2 & 3.06 & 1.06 & 100 & 23.5 & 1.95 & 1.31 \\
\hline
MJSMA\_F & 1 & 100 & \textbf{14.1} & \textbf{2.93} & 0.82 & 100 & 18.8 & 3.38 & 1.12 & 100 & 19.1 & 2.27 & 1.31 \\
\hline
\hline
JSMA+Z & 1 & 100 & 52.0 & 5.19 & \textbf{0.60} & 100 & 45.1 & 5.10 & 0.95 & 100 & 42.2 & 3.24 & \textbf{1.36} \\
\hline
JSMA-Z & 1 & 99.1 & 44.5 & 4.79 & 0.66 & 95.4 & 88.5 & 5.07 & \textbf{0.89} & 100 & 54.3 & 2.93 & 1.38 \\
\hline
NT-JSMA+Z & 1 & 100 & 20.0 & 3.47 & 1.14 & 99.9 & \textbf{19.6} & 3.41 & 1.29 & 99.9 & \textbf{22.6} & 2.50 & 1.61 \\
\hline
NT-JSMA-Z & 1 & 100 & 19.0 & 3.42 & 1.20 & 99.8 & 39.1 & \textbf{3.29} & 1.29 & 100 & 28.6 & \textbf{2.18} & 1.61 \\
\hline
M-JSMA\_Z & 1 & 100 & \textbf{15.3} & \textbf{3.06} & 1.01 & 99.9 & 22.0 & 3.69 & 1.25 & 100 & 25.7 & 2.61 & 1.41 \\
\hline
\hline
JSMA+F & 100 & 0.1 & 2.6 & 0.89 & 0.00 & 2.6 & 3.6 & 0.95 & 0.03 & 4.5 & 2.9 & 0.95 & 0.03 \\
\hline
JSMA-F & 100 & 0.1 & 2.1 & 0.97 & 0.00 & 2.6 & 4.0 & \textbf{0.91} & 0.04 & 5.0 & 3.0 & 0.65 & 0.03 \\
\hline
NT-JSMA+F & 100 & 0.1 & 2.4 & \textbf{0.85} & 0.00 & 2.7 & 3.6 & 0.95 & 0.03 & 4.6 & 2.9 & 0.96 & 0.03 \\
\hline
NT-JSMA-F & 100 & 0.1 & 2.1 & 1.01 & 0.00 & 2.8 & 4.0 & 0.93 & 0.03 & 5.0 & 3.0 & \textbf{0.65} & 0.03 \\
\hline
M-JSMA\_F & 100 & 0.1 & \textbf{2.0} & 0.97 & \textbf{0.00} & 2.6 & \textbf{3.3} & 1.10 & \textbf{0.02} & 4.8 & \textbf{2.8} & 0.79 & \textbf{0.03} \\
\hline
\hline
JSMA+Z & 100 & 100 & 37.1 & 4.43 & \textbf{0.01} & 100 & 32.5 & 4.40 & \textbf{0.02} & 100 & 46.8 & 3.48 & 0.04 \\
\hline
JSMA-Z & 100 & 98.7 & 42.4 & 4.55 & 0.01 & 95.6 & 58.0 & 3.86 & 0.04 & 100 & 59.8 & 3.21 & 0.05 \\
\hline
NT-JSMA+Z & 100 & 100 & 16.2 & 3.33 & 0.02 & 100 & \textbf{21.5} & 3.58 & 0.03 & 100 & \textbf{23.6} & 2.56 & 0.04 \\
\hline
NT-JSMA-Z & 100 & 100 & 19.8 & 3.52 & 0.02 & 100 & 39.8 & \textbf{3.26} & 0.05 & 100 & 27.3 & \textbf{2.13} & 0.04 \\
\hline
M-JSMA\_Z & 100 & 100 & \textbf{14.7} & \textbf{3.09} & 0.02 & 99.6 & 28.0 & 3.92 & 0.03 & 100 & 27.1 & 2.70 & \textbf{0.04} \\
\hline
\end{tabular}
\label{table:exp3}
\end{table}

Table~\ref{table:exp3} presents four sets of results that contrast softmax-layer attacks versus logit-layer attacks, and for the plainly-distilled $T=1$ and defensively-distilled $T=100$ temperatures. We begin by noting that the first block of results closely resemble statistics from Table~\ref{table:exp1}. This suggests that non-defensive distillation has minimal effects on adversarial attacks like JSMA.

Looking at the second block-row next, we observe that adversaries found by attacking the logit layers ($\{\pm Z\}$) consistently suffered from poorer perceptual similarities, as reflected by larger average $L_0$ and $L_2$ statistics.

Moving on, we see that all the $F$ attack variants failed to break defensively distilled models ($T=100$), which is consistent with reports by~\cite{DBLP:conf/sp/Carlini017} and~\cite{7546524}. Although there is a modification of JSMA proposed by~\cite{DBLP:journals/corr/CarliniW16} that can circumvent defensive distillation by dividing the logits of the distilled model by a temperature constant, we do not assume knowledge of the defense strategy used by the target model. However, in contrast to findings of~\cite{DBLP:conf/sp/Carlini017}, all $Z$ attack attempts were able to fool these distilled models.
Although the resulting statistics do not point to a single dominant variant, NT-JSMA$\pm$Z and M-JSMA\_Z both found adversaries with similarly small perceptual differences in comparably few iterations, while the targeted JSMA$\pm$Z trailed behind consistently.


\section{Conclusions}

We introduced Non-Targeted JSMA and Maximal JSMA as more flexible variants of the Jacobian-based Saliency Map Attack~\cite{DBLP:conf/eurosp/PapernotMJFCS16}, for finding adversarial examples both quickly and with limited perceptual differences.
Most notably, M-JSMA subsumes the need to specify the target class and the perturbation direction.
We empirically showed that M-JSMA consistently found high-quality adversaries among a variety of image datasets.
With this work, we hope to raise awareness of the ease of generating adversarial examples, and to develop better understandings of attacks so as to defend against them.

\bibliographystyle{IEEEtran}
\bibliography{mjsma.bib}

\end{document}